\documentclass{llncs}
\usepackage{isolatin1}
\usepackage{graphicx}

\newcommand{\DAE}{{\em Divide-and-Evolve }}

\begin{document}

\title{Divide-and-Evolve: a New Memetic Scheme for Domain-Independent Temporal Planning}

 \author{Marc Schoenauer$^1$, Pierre Savéant$^2$, Vincent Vidal$^3$}

 \institute{$^1$ Projet TAO, INRIA Futurs, LRI, Bt. 490, Université Paris Sud, 91405 Orsay, France\\
 $^2$  Thales Research \& Technology France, RD 128, F-91767 Palaiseau, France\\
 $^3$ CRIL \& Université d'Artois, rue de l'université - SP16, 62307 Lens, France\\
 }

\maketitle

\begin{abstract}
An original approach, termed \DAE is proposed
to hybridize Evolutionary Algorithms (EAs) with 
Operational Research (OR) methods in the domain of
Temporal Planning Problems (TPPs). Whereas standard
Memetic Algorithms use local search methods to improve the
evolutionary solutions, and thus fail when the local method stops
working on the complete problem, the \DAE  approach
splits the problem at hand into several, hopefully easier, 
sub-problems, and can thus
solve globally problems that are intractable when directly fed into
deterministic OR algorithms. 
But the most prominent advantage of the \DAE approach is that it 
immediately opens up an avenue for multi-objective optimization,
even though the OR method that is used is single-objective.
Proof of concept approach on the standard (single-objective) 
Zeno transportation benchmark
is given, and a small original multi-objective benchmark is proposed
in the same Zeno framework to assess the multi-objective capabilities
of the proposed methodology, a breakthrough in Temporal Planning.
\end{abstract}

\section{Introduction}

\emph{Artificial Intelligence Planning} is a form of general problem solving
task which focuses on problems that map into \emph{state models} that can be
defined by a state space $S$, an initial state $s_0 \subseteq S$, a set of goal states
$S_G \subseteq S$,
a set of actions $A(s)$ applicable in each state $S$, and a
transition function $f(a,s)=s'$ with $a\in A(s)$, and $s, s'\in S$. A solution to
this class of models is a sequence of applicable actions mapping the initial
state $s_0$ to a goal state that belongs to $S_G$.

An important class of problems is covered by Temporal Planning which extends 
classical planning by adding a duration to actions and by allowing concurrent actions in 
time \cite{Geffner-AAAI2002}.               
In addition, other metrics are usually needed for real-life problems to qualify a
good plan, for instance a cost or a risk criterion. A usual approach is
to aggregate the multiple criteria, but this relies on highly
problem-dependent  features and is not always meaningful. 
A better solution is to compute the set of optimal non-dominated solutions
 -- the so-called Pareto front.

Because of the high combinatorial complexity and the multi-objective
features of Temporal Planning Problems (TPPs), Evolutionary Algorithms are good general-purpose
candidate methods. 


However, there has been very few attempts to apply Evolutionary Algorithms to
planning problems and, as far as we know, not any to Temporal
Planning. Some approaches 
use a specific representation (e.g. dedicated
to the battlefield courses of action \cite{Goldberg-ECJ-1999}). 
Most of the domain-independent
approaches see a plan as a program and rely on Genetic Programming and
on the traditional blocks-world domain for experimentation (starting
with the Genetic Planner \cite{Spector-AAAI-94}). A more comprehensive
state of the art on Genetic Planning can be found in \cite{Morignot-2005}
where the authors experimented a variable length chromosome
representation. It is important to notice that all those works search
the space of (partial) plans.


It is also now well-known that Evolutionary Algorithms (EAs) can
rarely efficiently
solve Combinatorial Optimization Problems on their own, i.e. without
being hybridized, one way or another, with local search {\em ad hoc} techniques. 
The most successful of such hybridizations use Operational Research methods to 
locally improve any offspring that was born 
from EA variation operators (crossover and mutation): such algorithms
have been termed ``Memetic Algorithms'' 
or ``Genetic Local Search'' \cite{Merz:Freisleben99}. 
Those methods are now the heart of a whole research field,
as witnessed by the series of WOMA's (Workshops on Memetic Algorithms)
organized every year now, Journal Special
Issues 
and edited books \cite{MemeticBook:2005}. 

However, most memetic approaches are based on finding local
improvements of candidate solutions proposed by the evolutionary
search mechanism using dedicated local search methods that have to
tackle the complete  problem. In some combinatorial
domains such as Temporal Planning, this simply proves to be impossible 
when reaching some level of complexity.

This paper proposes {\em Divide-and-Evolve}, borrowing to the
Divide-and-Conquer paradigm  
for such situations: the problem at hand is sliced into a sequence of 
problems that are hopefully easier to solve by OR or other local
methods. The solution to the original problem is then obtained by a
concatenation of the solutions to the different sub-problems.





Next section presents an abstract formulation of the \DAE sche\-me, and
starting from its historical (and 
pedagogical) root, the TGV paradigm. Generic representation and
variation operators are also introduced.
Section \ref{planning} introduces an actual instantiation of the \DAE
scheme to TPPs. The formal framework of TPPs is
first introduced, then the TPP-specific issues for the \DAE
implementation are presented and discussed.
Section \ref{experiments} is devoted to experiments on the transportation 
Zeno benchmark for both single and multi-objective cases.
The last section opens a discussion highlighting the limitations of
the present work and giving hints about on-going and future work.

\section{The \DAE Paradigm}
\label{TGV-paradigm}

\subsection{The TGV metaphor}
\label{TGVMetaphor}
The \DAE strategy springs from a metaphor on the route
planning problem for the French high-speed train (TGV). The original problem
consists in computing the shortest route between two points of a
geographical landscape with strong bounds on the curvature and slope
of the trajectory.  
An evolutionary algorithm was designed \cite{RapportTGV92}
based on the fact that the only local 
search algorithm at hand was a greedy deterministic algorithm that 
could solve only very simple (i.e.  short distance) problems. The
evolutionary algorithm looks for a split of the global
route into small consecutive segments such that a local
search algorithm can easily find a route joining their extremities. 
Individuals represent sets of
intermediate train stations between the station of departure and the
terminus. The convergence toward a good solution was obtained with the
definition of appropriate variation and selection operators \cite{RapportTGV92}. 
Here, the state space is the surface on which the trajectory of the
train is defined.\\

\hspace{-0.5cm}{\bf Generalization}
Abstracted to Planning, the route is replaced by a sequence of actions
and the ``stations'' become intermediate states of the system. The
problem is thus divided into sub-problems and ``to be close'' becomes ``to
be easy to solve'' by some local algorithm ${\cal L}$.
The evolutionary algorithm plays the role of an oracle pointing at some 
imperative states worth to go trough. 

\subsection{Representation}
\label{representation}
The problem at hand is an abstract AI Planning problem as described in
the introduction.
The representation used by the evolutionary algorithm is a variable
length list of states: an individual is thus defined as
$(s_i)_{i\in [1,n]}$, where the length $n$ and all the states $s_i$ are
unknown and subject to evolution. States $s_0$ and $s_{n+1} \equiv s_G$ will
represent the initial state and the goal of the problem at hand, but
will not be encoded in the genotypes.
By reference to the original TGV paradigm, each of 
the states $s_i$ of an individual will be called a {\em station}.\\

\hspace{-0.5cm}{\bf Requirements}
The original TGV problem is purely topological with no temporal dimension
and reduces to a planning problem with a unique action: moving between
two points. The generalization to a given planning domain
requires to be able to: 

\begin{enumerate}
\item define a distance between two different states of the system, so
that $d(S, T)$ is somehow related to the difficulty for the
local algorithm ${\cal L}$ to find a plan mapping the initial state $S$
to the final state $T$;
\item generate a chronological sequence of virtual ``stations'',
i.e. intermediate states of the system, that are close to one another,
$s_i$ being close to $s_{i+1}$;
\item solve the resulting "easy" problems using the local algorithm ${\cal L}$;
\item ``glue'' the sub-plans into an overall plan of the problem at hand.
\end{enumerate}

\subsection{Variation operators}
\label{operators}
This section describes several variation operators that can be
defined for the general \DAE approach, independently of the
actual domain of application (e.g. TPPs, or the original TGV problem).\\

\hspace{-0.5cm}{\bf Crossover}
Crossover operators amounts to exchanging stations
between two individuals. Because of the sequential nature of the
fitness, it seems a good idea to try to preserve sequences of
stations, resulting in straightforward adaptations to variable-length
representation of the classical 1- or
2-point crossover operators.

Suppose you are recombining two
individuals $(s_i)_{1\leq n}$ and $(T_i)_{1\leq m}$. The 1-point crossover 
amounts to choosing one station in each individual, say $s_a$ and
${T_b}$, and exchanging the 
second part of the lists of stations, obtaining the two offspring
$(s_1, \ldots, s_a, T_{m+1}, \ldots T_b)$ and $(T_1, \ldots, T_b,
s_{n+1}, \ldots, s_n)$
(2-point crossover is easily implemented in a similar way). Note that
in both cases, the length of each offspring is likely to
differ from those of the parents.

The choice of the crossover points $s_a$ and $T_b$ can be either
uniform (as done in all the work presented here), or distance-based,
if some distance is available:
pick the first station $s_a$ randomly, and choose $T_b$ by e.g. a
tournament based on the distance with $s_a$ (this is on-going work).\\

\hspace{-0.5cm}{\bf Mutation}
Several mutation operators can be defined. Suppose individual
$(s_i)_{1\leq n}$ is being mutated:
\begin{itemize}
\item {\bf At the individual level}, the {\em Add} mutation simply
inserts a new station $s_{new}$ after a given station ($s_a$),
resulting in an $n+1$-long list, $(s_1, \ldots, 
s_a,$ $s_{new}, s_{a+1}, \ldots, s_n)$.
Its counterpart, the {\em Del} mutation, removes a station $s_a$ 
from the list. 

Several improvements on the pure uniform choice of $s_a$ 
can be added and are part of on-going work, too: in case the local
algorithm fails to successfully join all pairs of successive
stations, the last station that was successfully reached by the local
algorithm can be preferred for station $s_a$ (in both the {\em Add} and
{\em Del} mutations). If all partial problems are solved, the most
difficult one (e.g. in terms of number of backtracks) can be chosen.

\item {\bf At the station level}, the definition of each station can be
modified -- but this is problem-dependent. However, assuming there
exists a 
station-mutation operator $\mu_S$, it is easy to define the
individual-mutation $M_{\mu_S}$ that will 
simply call $\mu_S$ on each station $s_i$ with a user-defined
probability $p_{\mu_S}$. Examples of operators $\mu_S$ will be given
in section \ref{planning}, while simple Gaussian mutation of the
$(x,y)$ coordinates of a station were used for the original TGV
problem \cite{RapportTGV92}.
\end{itemize}

\section{Application to Temporal Planning}
\label{planning}
 
\subsection{Temporal planning problems}
\label{TPP}

Domain-Independent planners rely on the Planning Domain
Definition Language (PDDL) \cite{pddl}, inherited from the STRIPS 
model \cite{fikes:strips}, to represent a planning problem.
In particular, this language is used for a
competition\footnote{{\tt http://ipc.icaps-conference.org/}}
which is held every two years since 1998.
The language has been extended for representing Temporal Planning Problems in
PDDL2.1 \cite{Fox-JAIR-2003}. For the sake of simplicity, 
the temporal model is often simplified as explained below \cite{vidal:aaai04}.

A \emph{Temporal PDDL Operator} is a tuple $o=\langle pre(o), add(o), del(o),
dur(o)\rangle$ where $pre(o)$, $add(o)$ and $del(o)$ are sets of ground atoms that
respectively denote the preconditions, add effects and del effects of $o$, and
$dur(o)$ is a rational number that denotes the \emph{duration} of $o$. The
operators in a PDDL input can be described with variables, used in predicates
such as \texttt{(at ?plane ?city)}. 

A \emph{Temporal Planning Problem} is a tuple $P=\langle A, I, O, G\rangle$, where $A$ is
a set of atoms representing all the possible facts in a world situation, $I$ and
$G$ are two sets of atoms that respectively denote the initial state and the
problem goals, and $O$ is a set of ground PDDL operators.

As is common in Partial Order Causal Link (POCL) Planning \cite{weld:planning},
two dummy actions are also considered,  $Start$ and $End$ with zero
durations, the 
first with an empty precondition and effect $I$; the latter with precondition
$G$ and empty effects.  Two actions $a$ and $a'$ interfere when one deletes a
precondition or positive effect of the other. The simple model of time
in \cite{weld:time} defines a valid plan as a plan where interfering actions
do not overlap in time. In other words, it is assumed that the preconditions need to
hold until the end of the action, and that the effects also hold at the end and
cannot be deleted during the execution by a concurrent action.  

A {\em schedule\/} $P$ is a finite set of actions occurrences $\langle a_i,t_i\rangle$,
$i =1 , \ldots, n$, where $a_i$ is an action and $t_i$ is a non-negative integer
indicating the starting time of $a_i$ (its ending time is $t_i+dur(a_i)$).
$P$ must include the $Start$ and $End$ actions, the former with time tag $0$.
The same action (except for these two) can be executed more than once in $P$ if $a_i
= a_j$ for $i \not= j$. 
Two action occurrences $a_i$ and $a_j$
{\em overlap\/} in $P$ if one starts before the other ends; namely if
$[t_i,t_i+dur(a_i)] \cap [t_j,t_j+dur(a_j)]$ contains more than one time point.

A schedule $P$ is a {\em valid plan\/} iff interfering actions do not overlap in
$P$ and for every action occurrence $\langle a_i,t_i\rangle$ in $P$ its preconditions $p \in pre(a)$
are true at time $t_i$.  This condition is inductively defined as follows: $p$
is true at time $t=0$ iff $p \in I$, and $p$ is true at time $t > 0$ if either
$p$ is true at time $t-1$ and no action $a$ in $P$ ending at $t$ deletes $p$, or
some action $a'$ in $P$ ending at $t$ adds $p$.
The {\em makespan\/} of a plan $P$ is the time tag of the $End$
action. 

\subsection{CPT: an optimal temporal planner}
\label{CPT}
An optimal temporal planner computes valid plans with minimum
makespan. Even though an optimal planner was not mandatory (as
discussed in section \ref{discussion}), we have chosen {\em CPT}
\cite{vidal:aaai04}, a freely-available optimal temporal planner, for its 
temporal dimension and for its constraint-based approach which provide
a very useful data structure when it comes to gluing the partial
solutions (see section \ref{representation}).
Indeed, since in Temporal Planning actions can overlap in time, the
simple concatenation of sub-plans, though providing a feasible
solution, obviously might produce a plan that is not optimal with
respect to the total makespan, even if the sequence of actions is the
optimal sequence.
However, thanks to the causal links and order constraints
maintained by CPT, an improved global plan can be obtained by shifting
sub-plans as early as possible in a final state of the algorithm.





\subsection{Rationale for using \DAE for Temporal Planning}
\label{rationale}

The reasons for the failure of standard OR methods addressing TPPs
come from the exponential complexity of the number of possible
actions when the number of objects involved in the problem
increases. It is known for a long time that taking into account the
interactions between sub-goals can decrease the complexity of
finding a plan, in particular when these sub-goals are
independent \cite{korf:subgoals}. Moreover, computing an ideal
ordering on sub-goals is as difficult as finding a plan
(PSPACE-hard), as demonstrated in \cite{koehler:agenda}.
The basic idea when using the \DAE approach is that each
local sub-plan (``joining'' stations $s_i$ and $s_{i+1}$) should be
easier to find than the global plan (joining the station of departure
$s_0$ and the terminus $s_{n+1}$). This will be now demonstrated 
on the Zeno transportation benchmark (see {\tt http://ipc.icaps-conference.org/}). 


Table \ref{zeno14} illustrates the decomposition of a relatively difficult
problem in the Zeno domain (\texttt{zeno14} from IPC-3 benchmarks), a
transportation problem with 5 planes (\texttt{plane1} to \texttt{plane5}) and
10 persons (\texttt{person0} to \texttt{person9}) to travel among
10 cities (\texttt{city0} to \texttt{city9}).

Analyzing the optimal solution found by CPT-3 
it was easy to manually divide the optimal ``route'' of this solution
in the state space into four intermediate stations between the initial state and the goal.
It can be seen that very few moves (plane or person) occur between two 
consecutive stations (the ones in bold in each column of Table \ref{zeno14}).
Each sub-plan is easily found by CPT, with a maximum of
195 backtracks and 4.34 seconds of search time. 
It should be noted that most of the time spent by CPT is for pre-processing: this 
operation is actually repeated each time CPT is called, but could be factorized at 
almost no cost. 

Note that the final step of the process is the compression of the five
sub-plans (see section \ref{representation}): it is here performed in
0.02 seconds without any backtracking, and  the overall makespan of
the plan is 772, much 
less than the sum of the individual makespans of each sub-plan 
($2051$).

To summarize, the recomposed plan, with a makespan of 772, required 
a total running time of 254.38 seconds (including only 7.5s of pure
search) and 228 backtracks altogether, whereas a plan 
with the optimal makespan of 476 is found by CPT in 4,205 seconds and
606,405 backtracks. Section \ref{discussion} will discuss this issue.

\begin{table}[htb!]
  \caption{State Decomposition of the Zeno14 Instance. (The new location of moved objects appears in bold.)}
  \label{zeno14}
  \footnotesize
  \begin{tabular}{|c||c|c|c|c|c|c|}
    \cline{1-7}
\textbf{Objects}&\textbf{Init}&\textbf{Station 1}&\textbf{Station
2}&\textbf{Station 3}&\textbf{Station 4}&\textbf{Goal}\\
\textbf{}&(station 0)& & & & & (station 5)\\
    \cline{1-7}
plane 1 & city 5 & {\bf city 6} & city 6 & city 6 & city 6 & city 6 \\
plane 2 & city 2 & city 2 & {\bf city 3} & city 3 & city 3 & city 3 \\
plane 3 & city 4 & city 4 & city 4 & {\bf city 9} & city 9 & city 9 \\
plane 4 & city 8 & city 8 & city 8 & city 8 & {\bf city 5} & city 5 \\
plane 5 & city 9 & city 9 & city 9 & city 9 & city 9 & {\bf city 8} \\
person 1 & city 9 & city 9 & city 9 & city 9 & city 9 & city 9 \\
person 2 & city 1 & city 1 & city 1 & city 1 & city 1 & {\bf city 8} \\
person 3 & city 0 & city 0 & {\bf city 2} & city 2 & city 2 & city 2 \\
person 4 & city 9 & city 9 & city 9 & {\bf city 7} & city 7 & city 7 \\
person 5 & city 6 & city 6 & city 6 & city 6 & city 6 & {\bf city 1} \\
person 6 & city 0 & {\bf city 6} & city 6 & city 6 & city 6 & city 6 \\
person 7 & city 7 & city 7 & city 7 & city 7 & {\bf city 5} & city 5 \\
person 8 & city 6 & city 6 & city 6 & city 6 & city 6 & {\bf city 1} \\
person 9 & city 4 & city 4 & city 4 & city 4 & {\bf city 5} & city 5 \\
person 0 & city 7 & city 7 & city 7 & {\bf city 9} & city 9 & city 9 \\
\cline{1-7}
\multicolumn{2}{|c|}{Makespan} & 350 & 350 & 280 & 549 & 522 \\
\multicolumn{2}{|c|}{Backtracks} & 1 & 0 & 0 & 195 & 32 \\
\multicolumn{2}{|c|}{Search time} & 0.89 & 0.13 & 0.52 & 4.34 & 1.64 \\
\multicolumn{2}{|c|}{Total time} & 49.10 & 49.65 & 49.78 & 54.00 & 51.83 \\
\cline{1-7}
\multicolumn{2}{c|}{~} & \multicolumn{2}{|c|}{\textbf{Compression}} &
& \multicolumn{2}{|c|}{\textbf{Global Search}} \\
    \cline{1-4} \cline{6-7}
    \multicolumn{2}{|c|}{Makespan}  & \multicolumn{2}{|c|}{772} & & \multicolumn{2}{|c|}{476} \\
    \multicolumn{2}{|c|}{Backtracks}  & \multicolumn{2}{|c|}{0} & & \multicolumn{2}{|c|}{606,405} \\
    \multicolumn{2}{|c|}{Search time}  & \multicolumn{2}{|c|}{0.01} & & \multicolumn{2}{|c|}{4,155.41} \\
    \multicolumn{2}{|c|}{Total time}  & \multicolumn{2}{|c|}{0.02 (total : 254.38)} 
& & \multicolumn{2}{|c|}{4,205.40} \\
    \cline{1-4} \cline{6-7}
  \end{tabular}
\end{table}



\subsection{Description of the state space}
\label{TPPrepresentation}

\subsubsection{Non-temporal states}
A natural state space for TPPs, as described at the beginning of this
section, would be the actual space of all possible time-stamped states of the system. 
Obviously, the size of such a space is far too big and we simplified it 
by restricting the stations to non-temporal states.
However, even with this simplification, not all ``non-temporal'' states
can be considered in the description of the ``stations''.\\

\hspace{-0.5cm}{\bf Limiting the possible states}
First, the space of all possible states grows exponentially with 
the size of the problem. Second, not all states are consistent 
w.r.t. the planning domain. For instance, an object cannot be located
at two places at the same time in a transportation
problem -- and inferring such state invariants is
feasible but not trivial \cite{Fox-JAIR-1998}.
Note also that determining plan existence from a propositional STRIPS
description has been proved to be PSPACE-complete
\cite{bylander:complexity}.

A possible way to overcome this difficulty would be to rely on the
local algorithm to (rapidly) check the consistency of a given
situation, and to penalize unreachable stations. 
However, this would clearly be a waste of computational
resources.

On the other hand, introducing domain knowledge into EAs has been
known for long as the royal road toward success in Evolutionary Computation
\cite{Grefenstette_TSP87}. Hence, it seems a more promising approach to add
state invariants to the description of the state space in order to
remove the inconsistent states as much as possible. The good thing is that it is not
necessary to remove {\em all} inconsistent states since, in any case, the local
algorithm is there to help the EA to spot them -- inconsistent stations
will be given poor fitness, and will not survive next selection steps.
In particular, only state invariants involving a single predicate have
been implemented in the present work.

\subsection{Representation of stations}
\label{StationRepresentation}
It was hence decided to describe the stations using {\bf only the
predicates that are present in the goal} of the overall problem, and
to maintain the state invariants based on the semantics of
the problem.

A good example is given in Table \ref{zeno14}: the goal of this
benchmark instance is to move the persons and planes in cities
listed in the last column. No other predicate than the
corresponding {\tt (at objectN cityM)} predicates is present in the
goal. Through a user-supplied file, the algorithm is told that only
the {\tt at} predicates will be used to represent the stations, with
the syntactic restrictions that within a given station, the first
argument of an {\tt at} predicate can appear only once ({\tt at} is
said to be {\em exclusive} with respect to its first argument). The
state space that will be explored by the algorithm thus amounts to a
vector of 15 fluents (instantiated predicates) denoting that an item is 
located in a city (a column of table \ref{zeno14}).
In addition, the actual implementation of a station includes the
possibility to ``remove'' (in fact, comment out) a predicate of the list: the
corresponding object will not move during this sub-plan.\\

\hspace{-0.5cm}{\bf Distance}
The distance between two stations
should reflect the difficulty for the local
algorithm to find a plan joining them. 
At the moment, a purely
syntactic domain-independent distance is used: the number of
different predicates not yet 
reached. The difficulty can then be estimated by the number of
backtracks needed by the local algorithm. 
It is reasonable to assume that indeed most local problems where
only a few predicates need to be changed from the initial state to
the goal will be easy for the local algorithm - though this is
certainly not true in all cases.


\subsection{Representation-specific operators}
\label{variations}


\subsubsection{Initialization}
First, the number of stations is chosen uniformly in a user-supplied
interval. The user also enters a maximal distance $d_{max}$ between
stations.
A matrix is then built, similar to the
top lines of table \ref{zeno14}: each line corresponds to one of the goal
predicates, each column is a station. Only the first and last columns
(corresponding to initial state and goal) are filled with values. 
A number of ``moves'' is then
randomly added in the matrix, at most $d_{max}$ per column, and at
least one per line. Additional moves are then added according to
another user-supplied parameter, and without exceeding the 
$d_{max}$ limit per column. 
The matrix is then filled with values, starting from both
ends (init and goal), constrained
column-wise by the state invariants. 
A final sweep on all predicates comments out some of the predicates with a given
probability.\\

\hspace{-0.5cm}{\bf Station mutation}
Thanks to the simplified representation of the states (a vector of
fluents with a set of state invariants), it is straightforward to
modify one station randomly: with a given probability, a new value
for the non-exclusive arguments is chosen among the possible values
respecting all constraints (including the distance constraints with
previous and next stations).
In addition, each predicate might be commented out from the station 
with a given probability, like in the initialization phase. 

\section{First Experiments}
\label{experiments}

\subsection{Single objective optimization}
\label{singleObjective}

Our main playground to validate the \DAE approach is that of
transportation problems, and started with the {\tt zeno} domain as
described in section \ref{rationale}.
As can be seen in table \ref{zeno14}, the description of the stations
in {\tt zeno} domain involves a single predicate, {\tt at}, with 
two arguments. It is {\em exclusive} w.r.t. its first argument. Three
instances have been tried, called {\tt zeno10}, {\tt zeno12} and {\tt
zeno14}, from the simplest to the hardest.\\

\hspace{-0.5cm}{\bf Algorithmic settings}
The EA that was used for the first implementation of the \DAE paradigm
use standard algorithmic settings at the population level: a
$(10,70)-ES$ evolution engine (10 
parents give birth to 70 children, and the best 10 among the children
become the next parents), the children are created using 25\%
1-point crossover (see section \ref{operators})
and 75\% mutation (individual level), out of which 25\% are the {\em
Add} (resp. {\em Del}) generic mutations (section
\ref{operators}). The remaining 50\% of the mutations call the 
problem-specific station mutation. Within a station mutation, 
a predicate is randomly changed in 75\% of the cases and a predicate
is removed (resp. restored) in each of the remaining 12.5\% cases.
 (see section \ref{variations}).  Initialization is performed using
initial size in $[2,10]$, maximum distance of 3 and probability to
comment out a predicate is set to 0.1.
Note that no lengthy parameter tuning was performed for those proof-of-concept
experiments, and the above  values were decided based upon a very
limited set of initial experiments.\\

\hspace{-0.5cm}{\bf The fitness}
The target objective is here the total makespan of a plan -- assuming
that a global plan can be found, i.e. that all problems
$(s_i,s_{i+1})$ can be solved by the local algorithm.
In case one of the local problems could not be solved, the individual
is declared {\em infeasible} and is penalized in such a way that all
unfeasible individuals were worse than any feasible one. Moreover, this
penalty is proportional to the number of remaining stations after the
failure, in order to provide a nice slope of the fitness landscape
toward feasibility.
For feasible individuals, an average of the total makespan and the sum
of the makespans of all partial problems is used: when only the total
makespan is used, some individuals start bloating, without much 
consequence on the total makespan thanks to the final compression that
is performed by CPT, but nevertheless slowing down the whole run
because of all the useless repeated calls to CPT.\\

\hspace{-0.5cm}{\bf Preliminary results}
The simple {\tt zeno10} (resp. {\tt zeno12}) instance can be solved very
easily by CPT-2 alone, in less than 2s (resp. 125s), finding the
optimal plans with makespan 453 (resp. 549) using 154 (resp. 27560)
backtracks. 

For {\tt zeno10}, all runs found the optimal solution in the very
first generations (i.e. the initialization procedure always produced
a feasible individual that CPT could compress to the optimal
makespan. 
For {\tt zeno12}, all runs found a sub-optimal solution with makespan
between 789 and 1222. Note that this final solution was found after 3
to 5 generations, the algorithm being stuck to this solution
thereafter. The CPU time needed for 10 generations is around 5 hours.

A more interesting case is that of {\tt zeno14}.
First of all, it is worth mentioning that the present \DAE EA 
uses as local algorithm CPT version 2, and this version of CPT was
unable to find a solution to {\tt zeno14}: the results given in table
\ref{zeno14} have been obtained using the (yet experimental and not
usable from within the EA) new version of CPT. 
But whereas it proved unable to solve the full problem,
CPT-2 could nevertheless be used to solve the hopefully small instances
of {\tt zeno14} domain that were generated  by the \DAE approach --
though taking a huge amount of CPU time for that. Setting a limit on
the number of backtracks allowed for CPT was also mandatory to force
CPT not to explore the too complex cases that would have resulted in a
never-returning call.

However, a feasible individual was found  in each of the only 2
runs we could run -- one generation (70 evaluations) taking more than 10 hours.
In the first run, a feasible individual was found in the
initial population, with makespan 1958, and the best solution had a
makespan of 773. In the other run, the first feasible solution was found
at generation 3 -- but the algorithm never improved on that first
feasible individual (makespan 1356). 

Though disappointing with respect to the overall performances of the
algorithm, those results nevertheless witness for the fact that the
\DAE approach can indeed solve a problem that could not be solved by CPT
alone (remember that the version of CPT that was used in all
experiments is by far less efficient than the one used to solve {\tt
zeno14} in section \ref{rationale}, and was not able to solve {\tt
zeno14} at all.

\subsection{A multi-objective problem}
\label{multiObjective}
\subsubsection{Problem description}
In order to test the feasibility of the multi-objective approach based
on the \DAE paradigm, we extended the {\tt zeno} benchmark with an
additional criterion, that can be interpreted either as a cost, or as
a risk: in the former case, this additional objective is an additive
measure, whereas in the latter case (risk) the aggregation function is
the {\tt max} operator.

\begin{figure}
\begin{center}
\begin{tabular}{cc}
\includegraphics[width=6cm]{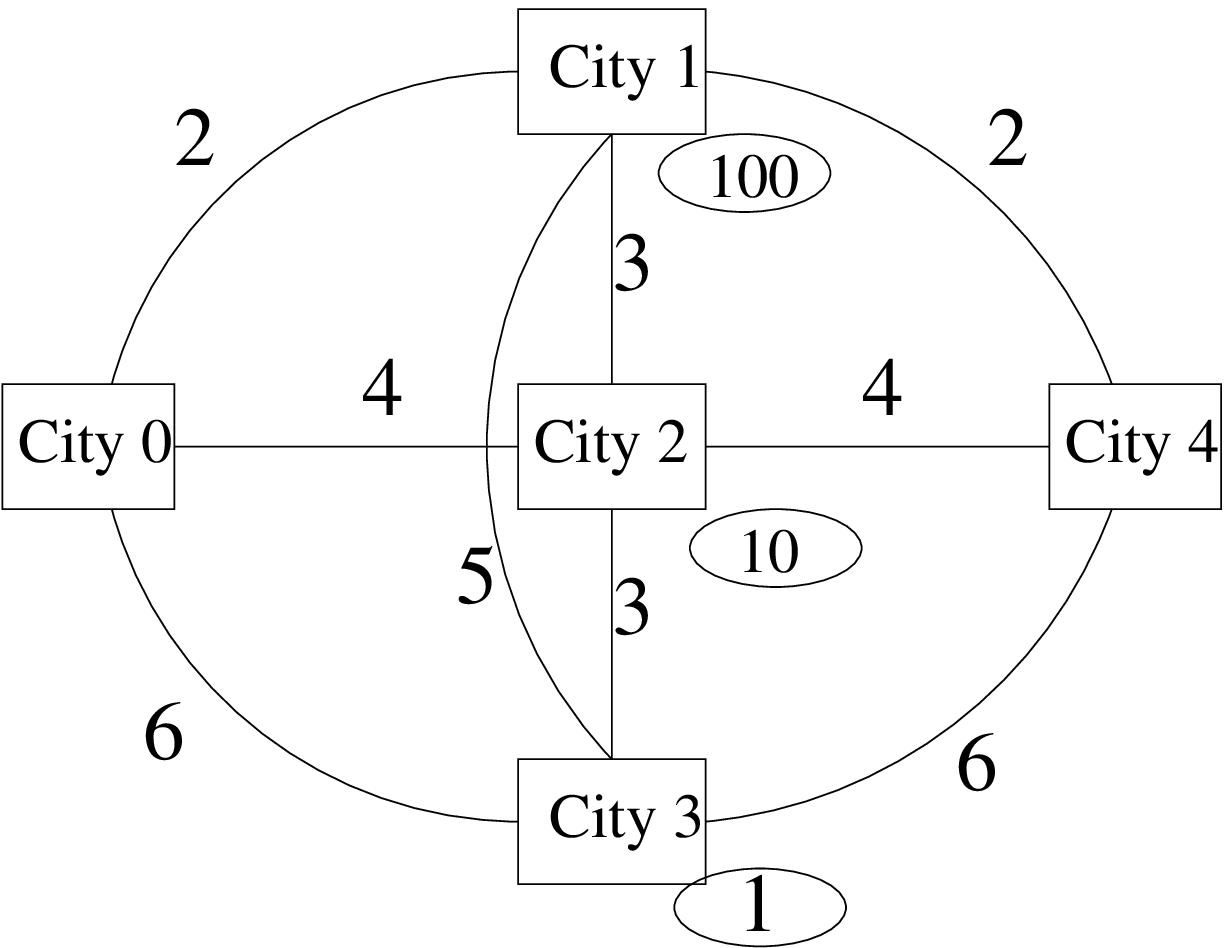} &
\includegraphics[width=6cm]{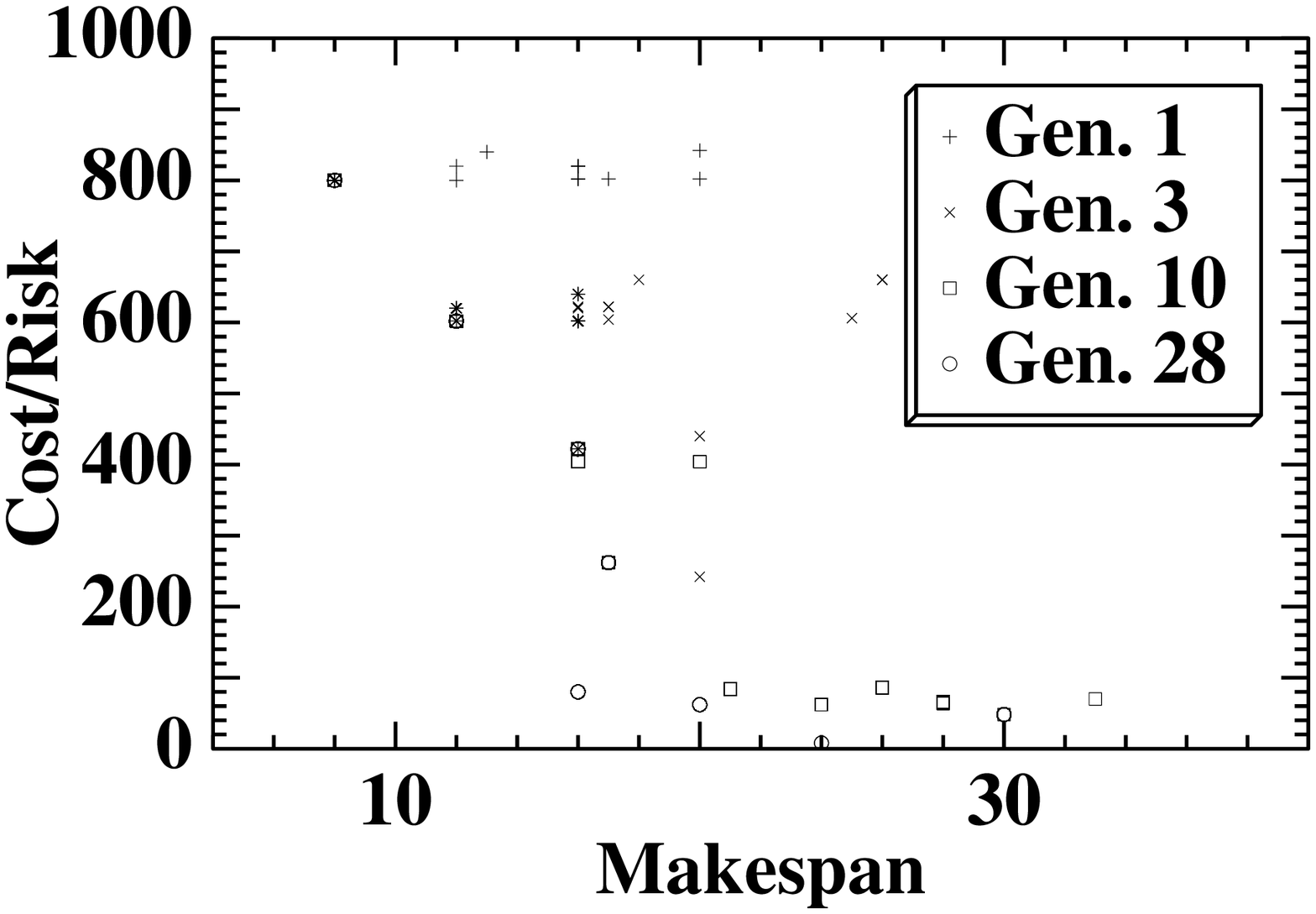}\\
\parbox{5.5cm}{a) The instance: Durations are attached to edges,
costs/risks are attached to cities (in gray circles). }
&
\parbox{5.5cm}{b) The population at different generations for a
successful run on the cost
(additive) instance of the zeno mini-problem of Figure 1-a.}
\end{tabular}
\caption{The multi-objective Zeno benchmark.}
\label{miniRisk} 
\end{center}
\end{figure}

The problem instance is shown in Figure \ref{miniRisk}: the only available
routes between cities are displayed as edges, only one transportation
method is available (plane), and the duration of the transport is shown on the
corresponding edge. Risks (or costs) are attached to the cities
(i.e., concern any transportation that either lands or takes off from that city).
In the initial state, the 3 persons and the 2 planes are in {\tt
  City~0},  and the goal is to transport them into {\tt City~4}.

As can be easily computed (though there is a little trick here), 
there are 3 remarkable
Pareto-optimal solutions, corresponding to traversing only one of the
3 middle cities. Going through {\tt City~1} is fast, but risky
(costly), whereas going through {\tt City~3} is slow and safe and cheap.

When all persons go
through respectively {\tt   City~1}, {\tt City~2} and {\tt   City~3},
the corresponding values of the makespans and costs in the additive
case are {\tt (8, 800)}, {\tt (16, 80)} and {\tt (24, 8)},
whereas they are, in the max case,  {\tt
  (8, 100)}, {\tt (16, 10)} and {\tt (24, 1)}.\\

\hspace{-0.5cm}{\bf Problem complexity}
It is easy to compute the number of possible virtual stations: 
 each one of the 3 persons can be in one of the 5 cities, or not
mentioned (absent predicate). Hence there are $3^6 = 729$
 possible combinations, and $729^n$ possible lists
 of length $n$. So even when $n$ is limited to 6, the size of the
search space is approx. $10^{17}$ \ldots\\

\hspace{-0.5cm}{\bf The algorithm}
The EA is based on the standard NSGA-II multi-objective EA \cite{Deb:NSGAII:PPSN2000}:
standard tournament selection of size 2 and deterministic replacement
among parents + offspring,
both based on the Pareto ranking and crowding distance selection;
a population size of 100 evolves during 30
generations. All other 
parameters were those used for the single objective case.\\

\hspace{-0.5cm}{\bf Fitnesses}
The problem has two objectives: one is the the total makespan (as in
the single-objective case), the other is either the {\bf risk} (aggregated
using the {\bf max} operator) or the {\bf cost} (an {\bf additive} objective).
Because the global risk only takes 3 values, there is no way to 
have any useful gradient information when used as fitness in the max case. 
However, even in the additive case, the same arguments than 
for the makespan apply (section \ref{singleObjective}), and hence, in
all cases, the  second objective is the
sum of the overall risk/cost and the average (not the sum) of the
values for all partial problems -- excluding from this average those
partial problems that have a null makespan (when the goal is already included 
in the initial state).\\

\hspace{-0.5cm}{\bf Results}
For the additive ({\bf cost}) case, the most difficult  Pareto optimum
(going through city 3 only) was
found 4 times out of 11 runs. However, the 2 other remarkable Pareto
optima, as well as  several other points in the Pareto front 
were also repeatedly found by all runs. Figure
\ref{miniRisk}-b shows different snapshots of the
population at different stages of the evolution for a typical
successful run: at first ('+'), all
individuals have a high cost (above 800); At generation 3 ('$\times$),
there exist 
individuals in the population that have cost less than 600; At
generation 10 (squares), many points have a cost less than 100. But
the optimal (24,8) solution is only found at generation 28 (circles).

The problem in the {\bf risk} context (the max case) proved to be, as
expected, slightly more difficult. All three Pareto optima (there
exist no other point of the true Pareto front in the max case) were
found only in 2 runs out of 11. However, all
runs found both the two other Pareto optima, as well as the slightly
sub-optimal solutions that goes only through city 3 but did not find
the little trick mentioned earlier, resulting in a (36,1)
solution.

In both cases, those results clearly validate the \DAE approach for
multi-objective TPPs --  remember that CPT has
no knowledge of the risk/cost in its optimization procedure - it only
aggregates the values 
a posteriori, after having computed its optimal plan based on the
makespan only -- hence the difficulty to find the 3rd Pareto optimum
going only through {\tt city3}.

\section{Discussion and Further Work}
\label{discussion}

A primary concern is the existence of a decomposition for any plan 
with optimal makespan. Because of the restriction of the
representation to the predicates that are in the goal, some states
become impossible to describe.
If one of these states is mandatory for all optimal plans, 
the evolutionary algorithm is unable to find the optimal solution.
In the {\tt zeno14} benchmark detailed in section \ref{rationale}, for
instance, one can see from the optimal solution that the {\tt in}
predicate should be taken   
into account when splitting the optimal solution, in order 
to be able to link a specific person to a specific plane. The main difficulty,
however, is to add the corresponding state invariant between {\tt
at} and {\tt in} (a person is either {\tt at} a location or {\tt in} a plane).
Future work will include state invariants involving pairs of 
predicates, to cope with such cases.
Along the same line, we will investigate whether it might be
possible to automatically infer some state invariants 
from the data structures maintained by CPT.


It is clear from the somehow disappointing results presented in section
\ref{singleObjective} that the search capabilities of the proposed algorithm
should be improved. But there is a lot of space for
improvements. First, and most immediate, the variation
operators could use some domain knowledge, as proposed in section
\ref{operators} -- even if this departs from ``pure'' evolutionary
blind search. 
Also, all parameters of the algorithm will be carefully fine-tuned.


Of course the \DAE scheme has to be experimented on more examples. 
The International Planning Competition provides many instances in 
several domains that are good candidates. Preliminary results on the
{\tt driver} problem showed very similar results that those reported
here on the {\tt zeno} domain. But other domains, such as 
the {\tt depot} domain, or many real-world domains, involve (at least)
2 predicates
in their goal descriptions (e.g., {\tt in} and {\tt on} for {\tt
depot}) . It is hence necessary to increase the range of allowed
expressions in the description of individuals.


Other improvements will result from the 
move to the new version of CPT, entirely rewritten in C.  It will be
possible to call CPT from 
within the EA, and hence to perform
all grounding, pre-processing and CSP representation only once: at the
moment, CPT is launched anew for each partial computation, and 
a quick look at table \ref{zeno14} shows that on {\tt
  zeno14} problem, for instance, the run-time per individual will
decrease from 250 to 55 seconds.
Though this will not {\em per se} improve the quality
of the results, it will allow us to tackle more complex problems than
even {\tt zeno14}. Along the same lines,  other planners, in
particular sub-optimal planners, will also be tried in lieu of CPT, as
maybe the \DAE approach 
could find optimal results using sub-optimal planners (as done in some
sense in the multi-objective case, see section \ref{multiObjective}).

A last but important remark about the results is that, at least in the
single objective case, the best solution found by the algorithm was
always found in the very early generations of the runs: it could be
that the simple splits of the problem into smaller sub-problems that
are done during the initialization are the main reasons for
the results. Detailed investigations will show whether or not an
Evolutionary Algorithm is useful in that context!

Nevertheless, we do believe that using Evolutionary Computation is
mandatory in order to solve multi-objective optimization problems, as
witnessed by the results of section 
\ref{multiObjective}, that are, to the
best of our knowledge, the first ever results of Pareto optimization
for TPPs.


\bibliographystyle{plain}
\bibliography{TGV}

\end{document}